# B-FIRE: Binning-Free Diffusion Implicit Neural Representation for Hyper-Accelerated Motion-Resolved MRI

Di Xu PhD[1], Hengjie Liu PhD[1], Yang Yang PhD[2], Mary Feng MD[1], Jin Ning PhD[3], Xin Miao PhD[4], Jessica E. Scholey PhD[1], Alexandra E. Hotca-cho MD[1], William C. Chen MD[1], Michael Ohliger PhD[2], Martina Descovich PhD[1], Huiming Dong PhD[1], Wensha Yang PhD[1], Ke Sheng PhD[1]

1 Radiation Oncology, University of California, San Francisco, California

2 Radiology and Biomedical Imaging, University of California, San Francisco, California

3 Siemens Medical Solutions USA, Inc., Cleveland, Ohio

4 Radiology at Children's Hospital Los Angeles, Keck School of Medicine, University of Southern California, Los Angeles, California

## Abstract

**Purpose**: Four-dimensional magnetic resonance image (4DMRI) is paramount for motion-resolved radiotherapy. Conventional 4DMRI relies on averaging data across multiple respiratory bins, resulting in blurred anatomy and misrepresentation of instantaneous motion dynamics. Moreover, the temporal resolution of 4DMRI is constrained by extended acquisition time to achieve data sufficiency. Reconstructing extremely undersampled data is needed to enable (near) real-time 4DMRI. We propose B-FIRE, a binning-free diffusion implicit neural representation (INR) framework for hyper-accelerated (>100x) 4DMRI reconstruction.

**Materials and Methods**: B-FIRE employs a CNN–INR encoder–decoder backbone optimized using diffusion with a comprehensive loss that enforces image-domain fidelity and frequency-aware constraints. Motion-binned image pairs were used as training references, while inference was performed on binning-free undersampled data. Experiments were conducted on a T1-weighted StarVIBE liver cohort (n=225), with accelerations ranging from 8 spokes/frame (RV8/46x) to RV1/375x. Quantitative evaluation was performed on a XCAT phantom cohort with modulated ground-truth motion. B-FIRE was compared against NuFFT, GRASP-CS, Cascade CNN, Re-Con-GAN, and CIRNet. Reconstruction fidelity (SSIM, PSNR, and RMSE), motion trajectory consistency, and inference latency were evaluated.

**Results**: B-FIRE consistently outperformed baselines across all accelerations, with the largest gain observed under extremely high acceleration ($56\% \uparrow$ SSIM to the next best at RV1). Reconstruction robustness was demonstrated across diverse breathing patterns on RV8, achieving superior-inferior mean-absolute-error of 0.5-1.3 mm. Dosimetric analysis indicated that B-FIRE enabled PTV margin (1-2 mm) reduction decreased OAR dose and unnecessary GTV dose escalation. Timing analysis showed that B-FIRE achieved 120 ms latency at $16 \times 16$ input size, placing it within the operational realm for real-time MRI.

**Conclusion**: B-FIRE is a binning-free framework enabling high-fidelity non-Cartesian hyper-accelerated 4DMRI. By preserving instantaneous internal motion while maintaining practical latency, B-FIRE provides an important step toward real-time volumetric MRI guidance, particularly for MR-guided radiotherapy that requires online intrafraction motion characterization.

## 1. Introduction

Fully sampled, high-quality magnetic resonance imaging (MRI) necessitates extended acquisition times as a direct consequence of the minimum k-space sampling density dictated by the Nyquist theorem[1,2], the inherently sequential nature of k-space data acquisition, and the pronounced sensitivity of MR signal encoding to physiological motion. These fundamental constraints impose coupled limitations on achievable spatial and temporal resolution, signal-to-noise ratio, and scan duration, thereby limiting the practical deployment of MRI in time-critical applications, including online MR-guided radiation therapy (MRgRT), where instantaneous 3D anatomical imaging would enable more accurate tumor and critical organ tracking, dose cumulation, and online adaptive RT[3]. These clinical needs have motivated extensive research in accelerated MR reconstruction using undersampled data.

Prior work has sought to improve imaging speed through optimized non-Cartesian k-space sampling trajectories[4] and parallel multi-receiver-coil encoding[5]. Correspondingly, accelerated MR reconstruction has been investigated along two primary directions: model-based approaches, exemplified by compressed sensing (CS)[6] and, more recently, data-driven deep learning (DL) approaches[7–11]. CS solves reconstruction as a regularized inverse problem that enforces predefined sparsity priors to compensate for undersampled k-space signals. For dynamic image series, the low-rankness in the temporal domain[12] and the sparseness of the deformation vector fields were exploited[13]. Yet, the achievable acceleration is fundamentally constrained by the finite spatial-temporal sparseness that can be theoretically exploited. Dictionary- and subspace-based methods extend CS by modelling temporal signals using predefined bases but remain limited by fixed representations under highly undersampled and irregular motion conditions[14–16]. DL approaches, on the other hand, reconstruct undersampled k-space data based on statistical learning. Existing DL methods mostly achieve similar image quality but with significantly faster reconstruction/inference speed[17].

Reconstruction strategies for motion-resolved MRI are commonly discussed regarding how k-space data is organized over respiratory cycles. Binning-free methods reconstruct each frame from a limited temporal acquisition window while leveraging spatiotemporal redundancies through appropriate priors. While theoretically offering a faithful representation of physiological dynamics, these methods are mostly restricted to 2D cine dynamic acquisitions, as it leads to severe sparsity in 3D acquisition, and reconstruction remains challenging even with advanced techniques. Conversely, motion-binned reconstruction mitigates this ill-posedness by pooling samples from multiple cycles into a small number of respiratory phases, effectively increasing sampling density at the cost of temporal fidelity due to intra-bin averaging and potential binning errors. As a pragmatic compromise of sparse 3D data, motion-binned frameworks remain the prevailing paradigm[18,19]. While limited efforts have demonstrated real-time 3D MRI, these have largely been confined to brain imaging, where motion is relatively constrained, and respiratory effects are minimal[20]. Therefore, pursuing higher accelerations with a binning-free approach is necessary to recover instantaneous motion information in respiration-driven imaging.

Given the physical limitations imposed by intrinsic imaging sparseness, the desired acceleration is likely to come from the statistical learning ability of DL methods.

Existing DL architectures are characterized by their methods of parameterization and convergence strategies. In terms of parameterization, convolutional neural networks (CNNs) and Transformers leverage efficient operators but impose discrete grid structures, restricting performance in non-Cartesian or highly accelerated acquisition[8,9]. Alternatively, Implicit Neural Representations (INRs) offer a resolution-independent, continuous framework consistent with MR physics, though this precision necessitates significant computational overhead[7]. In terms of optimization strategies, Generative Adversarial Networks (GANs)[21] facilitate fast convergence through adversarial learning but carry the risk of generating anatomically inconsistent hallucinations. Diffusion Probabilistic Models (DPMs)[22] mitigate these artifacts by employing likelihood-consistent denoising, yielding superior stability in ill-posed reconstruction tasks[11].

In the domain of discrete architectures, Schlemper et al.[8] and Xu et al.[9] utilized Cascade CNNs and Transformers to achieve acceleration factors of 11-fold (11x) and 9x, respectively. However, both approaches are constrained by grid-based operators that faltered under non-Cartesian sampling. To address this limitation, Xu et al.[7] employed a continuous INR-based parameterization to achieve up to 20x acceleration. However, it requires computationally intensive patient-specific optimization with slow convergence, and its practical applicability is further constrained when extending to higher-dimensional or more aggressively undersampled settings. Innovations in training paradigms have yielded mixed results: while the GAN-based Re-Con-GAN[23] struggled with data consistency beyond 10x acceleration[10], the diffusion-based CIRNet[11] achieved robust reconstruction up to 20x. Nevertheless, the scalability of CIRNet is hindered by high computational overhead. Consequently, no existing methodology—whether based on discrete, continuous, generative, or Markov-chain frameworks—successfully enables sub-second signal acquisition or binning-free reconstruction.

To bridge these methodological gaps and achieve *hyper-acceleration* (>100x) necessary for resolving instantaneous 3D anatomy, we introduce **B-FIRE** (Binning-Free diffusion Implicit neural REpresentation). This framework integrates a hybrid representation backbone—coupling a CNN encoder with an INR decoder—within DPM optimization paradigm. B-FIRE effectively synergizes the strengths of DPMs and INRs to recover intricate high-frequency details, while simultaneously employing the CNN encoder to project highly undersampled measurements into a compact latent space. Together with a task-aware loss objective that jointly enforces image-domain fidelity and spectral k-space consistency, this latent compression strategy significantly accelerates convergence and improves overall learning efficiency.

## 2. Materials and Methods

### 2.1 T1 StarVIBE Liver Data Cohort

The study was approved by the local Institutional Review Board at XXXX (#XXXX). A total of 225 patients (136 males and 89 females; average age of $65.76 \pm 12.1$) undergoing liver RT for primary hepatic malignancies or liver metastases were included. All patients received hepatobiliary contrast (gadoxeric acid; Eovist, Bayer; dosage determined upon patient weight per standard-of-care) and were scanned on a 3T MRI scanner (MAGNETOM Vida, Siemens Healthineers, Forchheim, Germany). A healthy volunteer was additionally scanned (no contrast injection) to provide controlled respiratory patterns—including deep-expiration breath-hold (DEBH), deep-inspiration breath-hold (DIBH), and free breathing—allowing verification that B-FIRE accurately reconstructs known and deliberately modulated motion trajectories.

A free-breathing T1-weighted volumetric golden angle stack-of-stars research sequence was used for signal acquisition. The scanning parameters were set as $TE/TR = 1.5/3\ ms$, matrix size (height × width) $nh \times nw = 288 \times 288$, field of view (FOV) $= 374 \times 374\ mm$, in-plane resolution $= 1.3 \times 1.3\ mm$, slice thickness $= 3\ mm$, radial views (RV) per partition $nViews = 3000$, number of channels $nC = 26$, number of slices $nZ = 64 - 75$, acquisition time $= 7 - 10\ min$ (average acquisition time $T_{acq}$/RV $= 160\ ms$). The pulse sequence ran continuously over multiple respiratory cycles. Only regular breathers (225 patients) were included with breathing regularity determined using self-gating signal waveform[24].

The training: test patient size $= 200{:}\,25\ \approx 9{:}\,1$(random split). Motion-binned training ground truths (GTs) were reconstructed using nonuniform fast Fourier transform (NuFFT) with amplitude-based motion binning[25] (8 respiratory phases) applied to the entire RV3000 and were treated as fully sampled reference images (i.e., based on Nyquist theorem, fully sampled images RV≥ $nw \times \frac{\pi}{2}$, resulting in 452 spokes for matrix size $= 288 \times 288$. $\frac{RV3000}{8} = RV375$ is close to RV452 [within ~17% of this threshold], supporting near-complete sampling). The paired motion-binned inputs were retrospectively undersampled by keeping the first 8 (46x, RV8, $T_{acq,RV8} = 1280\ ms$), 5 (75x, RV5, $T_{acq,RV5} = 800\ ms$), 3 (125x, RV3, $T_{acq,RV3} = 480\ ms$), 2 (188x, RV2, $T_{acq,RV2} = 320\ ms$) and 1 (375x, RV1, $T_{acq,RV1} = 160\ ms$) spoke(s) per motion bin of RV3000, respectively.

For evaluation, a fully controlled digital phantom framework based on the extended cardiac-torso (XCAT) model[26] was adopted to enable quantitative assessment of binning-free reconstruction with known GT motion. All patients in the test split were used to generate subject-specific phantoms, with static volumes reconstructed via NuFFT from RV3000 serving as anatomical reference. Four-dimensional (4D) sequences were then synthesized with the XCAT deformable motion model[26], simulating six different respiratory liver motion patterns per patient ($25 \times 6 = 150$ scans in total), including periodic breathing (188 temporal phases at the RV acquisition interval, 4 $s$ /breathing cycle, and SI [near diaphragm]/AP/ML amplitudes of 12/3/1 $mm$), amplitude variation (~20–30% SI amplitude), baseline drift (~20% gradual SI amplitude drift), breath rate variability (cycle duration $\in [3,4]\ s$) and breath-hold (15 $s$; DIBH and DEBH) conditions.

For each temporal phase, the image volume was first modulated by synthetic multi-coil sensitivity profiles to produce coil-encoded complex images. K-space data was then generated from the complex images using a NuFFT-based forward operator with a stack-of-stars trajectory that replicates the original scanner data format. A fixed number of spokes per phase (RV8-RV1) was assigned following a global continuous golden-angle rotation across all temporal phases, simulating a binning-free acquisition scheme. The resulting raw k-space data was subsequently reconstructed using adjoint NuFFT by grouping continuous spokes (RV8-RV1) per phase in acquisition order to produce binning-free undersampled image-domain inputs.

Both the training and test undersampled inputs were processed in a 2.5D configuration ($z \times h \times w = 3 \times 288 \times 288$), where each sample consists of a stack formed by a central axial slice together with its immediately adjacent superior and inferior slices to provide through-plane contextual information while maintaining computational efficiency.

## 2.2 Conditional Diffusion Probabilistic Modelling Process

B-FIRE is an end-to-end framework, as illustrated in Fig. 1, with the inference process of DPM shown in **Fig. 1 (a)**. Given an under- and fully sampled image pair $(\boldsymbol{x}_i, \boldsymbol{y}_i)$, B-FIRE aims to learn a parametric approximation of data distribution $p(\boldsymbol{y}|\boldsymbol{x})$ via a fixed Markov chain of length $T$. The forward Markovian diffusion process $q$ with incremental addition of Gaussian noise $\mathcal{N}$ was defined as equations (1-2):

$$q(\boldsymbol{y}_{1:T}|\boldsymbol{y}_0) = \prod_{t=1}^{T} q(\boldsymbol{y}_t|\boldsymbol{y}_{t-1}) \quad (1),$$

$$q(\boldsymbol{y}_t|\boldsymbol{y}_{t-1}) = \mathcal{N}(\boldsymbol{y}_t|\sqrt{1-\beta_t}\boldsymbol{y}_{t-1}, \beta_t \boldsymbol{I}) \quad (2),$$

where $\beta_t \in (0,1)$ is the variance of $\mathcal{N}$ in $T$ iterations. The distribution of $q(\boldsymbol{y}_t|\boldsymbol{y}_0)$ can be represented as equation (3):

$$q(\boldsymbol{y}_t|\boldsymbol{y}_0) = \mathcal{N}(\boldsymbol{y}_t|\sqrt{\gamma_t}\boldsymbol{y}_0, (1-\gamma_t)\boldsymbol{I}) \quad (3),$$

where $\gamma_t = \prod_{i=1}^{t}(1-\beta_i)$. At inference stage, B-FIRE conducted a conditional reverse Markovian process $p_\theta(\boldsymbol{y}_{t-1}|\boldsymbol{y}_t, \boldsymbol{x})$ as equations (4-6):

$$p_\theta(\boldsymbol{y}_{0:T}|\boldsymbol{x}) = p(\boldsymbol{y}_T)\prod_{t=1}^{T} p_\theta(\boldsymbol{y}_{t-1}|\boldsymbol{y}_t,\boldsymbol{x}) \quad (4),$$

$$p(\boldsymbol{y}_T) = \mathcal{N}(\boldsymbol{y}_T|0,\boldsymbol{I}) \quad (5),$$

$$p_\theta(\boldsymbol{y}_{t-1}|\boldsymbol{y}_t,\boldsymbol{x}) = \mathcal{N}(\boldsymbol{y}_{t-1}|[\boldsymbol{\mu}_\theta \cup \boldsymbol{f}_\theta \rightarrow \boldsymbol{g}_\theta](\boldsymbol{x},\boldsymbol{y}_t,t),\sigma_t^2\boldsymbol{I}) \quad (6),$$

where $\boldsymbol{\mu}_\theta \cup \boldsymbol{r}_\theta \rightarrow \boldsymbol{g}_\theta$ denotes the representation backbone consists of a CNN encoder consisting of U-Net $\boldsymbol{\mu}_\theta$ and ResNet[27] $\boldsymbol{f}_\theta$ and INR decoder $\boldsymbol{g}_\theta$ as shown in Fig. 1(b-c) with $\boldsymbol{u}^{(i)}$, $\boldsymbol{f}^{(i)}$ and $\boldsymbol{g}^{(i)}$ represents the number of layers in the network structure.

### 2.3 Conditioning Mechanism

**Convolutional Reconstruction Encoder**: Following the design of IDM[28], a CNN structure, consisting of EDSR[29], U-Net, and ResNet, was employed as the conditioning encoder to extract features into multiple resolutions from input undersampled images (**Fig. 1(b)**). Specifically, EDSR established the initial processing with an output of $\boldsymbol{f}^0$. Next, $\boldsymbol{f}^0$ and $\boldsymbol{y}_t$ were channel-wise concatenated and fed into the U-Net encoder while $\boldsymbol{f}^0$ was fed into ResNet encoder to form the two cascade branches for preliminary conditional guidance. Lastly, the output from the corresponding U-Net and ResNet encoder layers, $\boldsymbol{f}^{(i)}$ and $\boldsymbol{u}^{(i)}$, was channel-wise concatenated and then sent to the Leaky ReLU activation function $\mathcal{A}$ to form the final encoded feature map $\boldsymbol{h}^{(i)}$ as equation (7):

$$\boldsymbol{h}^{(i)} = \mathcal{A}\left(\boldsymbol{f}^{(i)},\boldsymbol{u}^{(i)}\right) \quad (7).$$

**Implicit Neural Representation Embedded Reconstruction Decoder**: As shown in **Fig. 1(c)**, INR was included at the decoder stage to parameterize features with a continuous representation. Specifically, we concatenated multiple coordinate-based multi-layer perceptrons (MLPs) as the up-sampling operations $\boldsymbol{D} = \{D^{(1)}, \dots, D^{(N)}\}$ to parameterize the INRs. $\boldsymbol{D}$ assumed continuous coordinate systems, $\boldsymbol{c} = \{\boldsymbol{c}^{(1)}, \dots, \boldsymbol{c}^{(N)}\}$, that represents its corresponding input dimension of $\boldsymbol{h}$, respectively. Given the features $\boldsymbol{h}^{(i+1)}$ and its associated coordinates $\boldsymbol{c}^{(i+1)}$, we formulated the INR process $\boldsymbol{g}$ as equation (8):

$$\boldsymbol{g}^{(i)} = D_i\left(\widehat{\boldsymbol{h}}^{(i+1)}, \boldsymbol{c}^{(i)} - \widehat{\boldsymbol{c}}^{(i+1)}\right) \quad (8),$$

where $D_i$ is a 2-layer MLP with 256 hidden neurons, and $\widehat{\boldsymbol{h}}^{(i+1)}$ and $\widehat{\boldsymbol{c}}^{(i+1)}$ were interpolated using the nearest Euclidean distance from $\boldsymbol{h}^{(i+1)}$ and $\boldsymbol{c}^{(i+1)}$ in the $(i+1)$-th depth, correspondingly.

## 2.4 Optimization

B-FIRE aims to recover the target image $\boldsymbol{y}_0$ through a sequence of denoising reconstruction steps. The process of restoring a target image $\boldsymbol{y}_0$ from an undersampled noisy image $\widehat{\boldsymbol{y}}_t$ is as equation (9):

$$\widehat{\boldsymbol{y}}_t = \sqrt{\gamma_t}\boldsymbol{y}_0 + \sqrt{1-\gamma_t}\boldsymbol{\epsilon} \quad (9),$$

which is equivalent to optimizing a noise model $\boldsymbol{\epsilon}_\theta$ that approximates the injected noise $\boldsymbol{\epsilon}$, thereby enabling progressive refinement towards the target image $\boldsymbol{y}_0$.

To achieve stable and accurate reconstruction under extreme undersampling, the optimization of B-FIRE combines diffusion-based noise modelling with spectral- and structure-aware reconstruction constraints. Accordingly, the diffusion component was trained using a L1-based denoising objective as equation (10):

$$\mathcal{L}_{diff} = \mathbb{E}_{(x,y)}\mathbb{E}_{\epsilon,\gamma_t,t}||\boldsymbol{\epsilon} - \boldsymbol{\epsilon}_\theta(\boldsymbol{x}, t, \widehat{\boldsymbol{y}}_t, \gamma_t)||_1^1 \quad (10),$$

where $\boldsymbol{\epsilon} \in \mathcal{N}(0, \boldsymbol{I})$, $t \in [1, \dots, T]$ and $(\boldsymbol{x}, \boldsymbol{y})$ was sampled from the training set of undersampled and fully sampled image pairs.

Additionally, a spectral attention k-loss was introduced to explicitly regulate reconstruction errors in the Fourier domain. In specific, pseudo-Cartesian-k-space representations were obtained by applying a 2D Fourier transform $\mathcal{F}$ to the predicted and reference magnitude images, and their discrepancy was penalized using a frequency-weighted L1 loss as Equation (11):

$$\mathcal{L}_{spec} = \mathbb{E}_{\boldsymbol{y}_0}||W_{HF} \odot |\mathcal{F}(\widehat{\boldsymbol{y}}_0) - \mathcal{F}(\boldsymbol{y}_0)|||_1^1 \quad (11),$$

where $W_{HF}$ was constructed using a radial weighting function over normalized k-space radius to emphasize high-frequency components that are critical for preserving edges and fine anatomical details under aggressive undersampling.

To further ensure image fidelity and structural consistency, B-FIRE incorporated an image-domain L1 loss as equation (12):

$$\mathcal{L}_{strc} = \mathbb{E}_{\boldsymbol{y}_0}||\widehat{\boldsymbol{y}}_0 - \boldsymbol{y}_0||_1^1 \quad (12).$$

Together with an edge-preserving L1 loss as equation (13):

$$\mathcal{L}_{edge} = \mathbb{E}_{\boldsymbol{y}_0}||\nabla\widehat{\boldsymbol{y}}_0 - \nabla\boldsymbol{y}_0||_1^1 \quad (13).$$

Where $\nabla$ denotes the spatial gradient operator computed using first-order finite differences along the in-plane directions. Collectively, the final loss objective was formed as equation (14):

$$\mathcal{L} = \lambda_{diff}\mathcal{L}_{diff} + \lambda_{spec}\mathcal{L}_{spec} + \lambda_{strc}\mathcal{L}_{strc} + \lambda_{edge}\mathcal{L}_{edge} \quad (14),$$

where $\lambda_{diff}$ , $\lambda_{spec}$ , $\lambda_{strc}$ and $\lambda_{edge}$ are the weighting hyperparameters for each corresponding loss terms that balance denoising accuracy, spectral fidelity structural and edge preservation. $\lambda_{diff}$ and $\lambda_{spec}$ were set as 2 while $\lambda_{strc}$ and $\lambda_{edge}$ were set as 1 in the current work (The loss weights were empirically selected; reconstruction performance was observed to be stable under moderate variations).

## 2.5 Training and Inference

To improve the robustness across different acquisition regimes, B-FIRE was trained using a hybrid undersampling strategy that jointly incorporates multiple acceleration factors (46x-375x). During Training, diffusion time steps were randomly sampled rather than sequentially unrolled, allowing efficient training optimization without requiring full diffusion processes.

At the inference stage, the trained B-FIRE was directly applied to process binning-free undersampled inputs, and reconstruction was performed through iterative denoising with a fixed number of diffusion time steps ($T = 1500$). The training-inference strategy allows B-FIRE to generalize from binned supervision to binning-free inference while maintaining robustness under extreme acceleration.

## 2.6 Benchmark Methods and Evaluation

Five representative reconstruction methods are selected. A direct NuFFT[30] reconstruction was used as a classical reference, while GRASP-CS[31] was selected as a conventional model-based iterative approach. Cascade CNN[8] was adopted as an unrolled CNN architecture. Re-Con-GAN[10] and CIRNet[11] were included as more recent approaches for highly undersampled 4DMRI, leveraging adversarial training and diffusion-based modelling, respectively. NuFFT and GRASP-CS were implemented on the Siemens ICE platform, while Cascade CNN, Re-Con-GAN and CIRNet were implemented in-house due to the absence of publicly available source code.

As defined in equations (15)-(18), reconstruction performance was evaluated on binning-free reconstruction using root mean squared error (RMSE), peak signal to noise ratio (PSNR), structural similarity index measurement (SSIM), and detection consistency index (DCI).

DCI quantifies the patient-wise proportion of phases where gross tumor volume (GTV) is successfully localized. All GTVs were delineated by experienced radiation oncologists on motion-averaged MRI in accordance with clinical practice. The dynamic positions of GTV were then determined by propagating the GTV centroid based on GT motion trajectory. Localization was assessed using a prompt-driven object detection model[32,33] initialized with a user-selected coordinate $S \in R^3$ within the GTV $G$ with successful detection defined as the predicted bounding box $B$ enclosing the GTV.

$$RMSE = \sqrt{\frac{\sum_{i=1}^{N} ||\boldsymbol{y}_i - \widehat{\boldsymbol{y}}_i||_2^2}{N}} \quad (15),$$

$$PSNR = 20 \cdot log_{10}\left(\frac{MAX_I}{RMSE}\right) \quad (16),$$

$$SSIM = \frac{(2\mu_{\boldsymbol{y}_i}\mu_{\widehat{\boldsymbol{y}}_i} + C_1)(2\sigma_{\boldsymbol{y}_i\widehat{\boldsymbol{y}}_i} + C_2)}{(\mu_{\boldsymbol{y}_i}^2 + \mu_{\widehat{\boldsymbol{y}}_i}^2 + C_1)(\sigma_{\boldsymbol{y}_i}^2 + \sigma_{\widehat{\boldsymbol{y}}_i}^2 + C_2)} \quad (17),$$

$$DCI = \frac{1}{N_P}\sum_{i=1}^{N_P}[B_i \sim G_i] \quad (18),$$

where $MAX_I$ is the maximal possible pixel value in a matrix, $\mu$ and $\sigma$ are the local mean and variance, $C_1 = (k_1 L)^2$ and $C_2 = (k_2 L)^2$ are stability constants with $k_1 = 0.01$, $k_2 = 0.03$, $L$ being the dynamic range of pixel values and $N_P$ is the patient-wise total number of motion frames.

## 2.7 Implementation Details

B-FIRE is trained end-to-end using a two-stage strategy with total number of trainable parameters of 94.92M. In the first stage, the model was trained for 1 million iterations using a fixed acceleration factor of 46x. In the second stage, training was continued for 0.5 million iterations, during which undersampled inputs were randomly selected from acceleration factors of 46x - 375x with a uniform probabilistic distribution.

Training was performed using the Adam optimizer with a dropout rate of 0.2. A fixed learning rate of $1e-4$ was used in the first stage and $2e-5$ in the second stage. All experiments were carried out on a $4\ \times$RTX A6000 GPU cluster (48 GB per GPU) with a training batch size of $4 \times 128$.

Inference was conducted on a $2 \times$ H200 GPU cluster (141 GB per GPU) to evaluate reconstruction efficiency under high-performance computing settings. Mixed-precision inference (FP16/FP8 where applicable) was employed to reduce computational latency. Inference time was benchmarked across different input patch sizes to assess scalability and real-time feasibility.

# 3. Experiments and Results

## 3.1 Reconstruction Quality Assessment

**Fig. 2** and **Tab. 1** summarize evaluation on the XCAT test cohort with known GT motion. **Fig. 2** shows representative B-FIRE reconstructions at 46x across multiple breathing patterns, while **Tab. 1** reports statistical comparisons against baselines over a range of acceleration factors.

**Fig. 2** demonstrates that B-FIRE robustly recovers respiratory motion across diverse patterns, including periodic breathing, amplitude variation, rate variability, baseline drift, DIBH, and DEBH. Across all scenarios (**Fig. 2 [a–f]**), predicted superior–inferior (SI) liver dome motion closely agrees with GT, with moderate phase lag and amplitude distortion, and slightly increased deviations under challenging conditions (e.g., breath-hold and baseline drift transitions). The SI mean absolute error (MAE) remains low (0.52–1.30 $mm$), indicating high fidelity even under irregular motion.

**Tab. 1** quantitatively evaluates reconstruction fidelity across accelerations from 46x to 375x for B-FIRE and baselines. B-FIRE consistently achieves the highest SSIM and PSNR and lowest RMSE, indicating superior structural preservation under extreme undersampling. At 46x, B-FIRE attains SSIM 0.85, outperforming CIRNet (0.68), Re-Con-GAN (0.64), Cascade CNN (0.60), CS (0.51), and NuFFT (0.48). B-FIRE also maintains consistently high DCI, supporting reliable target localization across continuous motion phases.

## 3.2 Comprehensive Motion Trajectory Analysis

**Fig. 3-4** evaluate B-FIRE binning-free reconstruction for recovering physiologically meaningful respiratory motion with two complementary objectives: (1) validating fidelity using a healthy volunteer with modulated breathing; (2) characterizing subtle hepatic dynamics suppressed by conventional motion-binned reconstruction.

In **Fig. 3 (a–c),** binning-free liver dome SI motion from a healthy volunteer shows distinct, temporally coherent patterns for DIBH, DEBH, and free breathing (two DIBH and two DEBH cycles acquired in one continuous scan, each held as tolerated with brief free-breathing intervals). DIBH exhibits greater residual fluctuation than DEBH (**Fig. 3 [a]**), consistent with respiratory biomechanics: deep inspiration requires active diaphragmatic engagement with higher variability, whereas deep expiration reflects a more stable relaxed state[34]. In **Fig. 3 (c)**, image-derived trajectories align in timing and trend with independently measured k-space-center SI projections, supporting physical plausibility. Unlike motion-binned reconstruction, which reduces motion to discrete synthetic phases, the binning-free approach preserves true physiological motion with larger excursions. In **Fig. 3 (f–g)**, liver dome and GTV-centroid trajectories in a representative patient show coherent respiratory modulation with increased SI excursion relative to motion-binned references, consistent with reduced temporal averaging.

Beyond validation, binning-free reconstruction enables analysis of subtle internal and external motion (**Fig. 3 [a–c, f–h]**) and their relationships (**Fig. 3 [e, j]**). External surface motion follows similar respiratory transitions but differs in smoothness and amplitude from the internal liver dome, indicating related but non-identical motion content. Pairwise scatter plots show weak-to-moderate, subject-specific correlations among SI-projected k-space center, internal, and external signals, reflecting both linear and nonlinear components.

**Fig. 4** illustrates the impact of increasing temporal resolution from RV8 to RV1. Despite greater reconstruction challenges at higher acceleration, SI motion trajectories from both image space and k-space centers remain detectable and temporally aligned. Panel (l) shows coherent liver dome motion across resolutions, with broadly consistent timing and envelopes. A clear trade-off is observed: lower resolutions yield smoother, dampened trajectories due to temporal averaging, while higher resolutions capture sharper fluctuations and larger amplitudes. Although high-frequency variation at extreme acceleration suggests some noise sensitivity, preserved global trends and phase alignment confirm reliable motion detection under extreme undersampling.

### 3.3 Dosimetric Impact of Planning Target Volume Margin Selection

To evaluate clinical implications of improved motion tracking, we assessed dosimetric impact of planning target volume (PTV) margins (1, 2, and 3 mm GTV expansion; **Fig. 5**). Margins of 1–2 $mm$ reflect B-FIRE reconstruction uncertainty (SI MAE: 0.52–1.30 $mm$), while 3 $mm$ serves as a reference consistent with the smallest reported margins in clinical MRgRT systems (ViewRay MRIdian and Elekta Unity)[35,36].

**Fig. 5 (a–b)** shows dosimetric differences across margin strategies. Increasing PTV margins broadens the prescription isodose region and elevates GTV dose due to normalization to PTV coverage, while also progressively increasing organs-at-risk (OAR) dose from 1 to 3 $mm$.

## 4. Discussion

We present B-FIRE (Binning-Free diffusion Implicit neural REpresentation) for hyper-accelerated, binning-free, motion-resolved non-Cartesian MRI reconstruction. It integrates a CNN encoder and INR decoder within a diffusion framework, with image- and frequency-domain consistency constraints, enabling robust binning-free recovery from RV8 to RV1. While hallucination is a concern at such extreme acceleration, B-FIRE mitigates this via data-conditioned diffusion, continuous INR modeling, and frequency-aware constraints that enforce structural consistency with measured signals. This framework enables recovery of instantaneous 3D anatomy over a large abdominal FOV without temporal data sharing, overcoming limitations of 2D or motion-averaged representations and enabling real-time 3D visualization of highly mobile anatomy.

Conventional motion-binned MR assumes respiratory repeatability, fitting multi-cycle data into a single synthetic cycle of discrete phases. While this mitigates undersampling for large-FOV 3D reconstruction and supports radiotherapy tasks (e.g., tumor/OAR motion quantification, internal target volume delineation, motion management), it suppresses non-periodic and cycle-to-cycle variations. In contrast, B-FIRE reveals larger, more nuanced continuous internal motion, providing clinically valuable information for motion-aware imaging and intervention (e.g., MRgRT), where accurate characterization of instantaneous intrafraction motion is critical[34,37,38].

Clinical MRgRT systems (e.g., ViewRay MRIdian, Elekta Unity) primarily use 2D cine MRI, which cannot capture through-plane motion and may misrepresent true 3D dynamics[39–41].

Volumetric real-time guidance is critical for upper-abdominal RT, where motion and deformation near dose-limiting gastrointestinal (GI) organs (e.g., duodenum, stomach, small bowel) constrain dose escalation and increase toxicity risk[42]. In principle, real-time 3D MRgRT can reduce motion margins via accurate 3D GTV localization, improving OAR sparing and reducing toxicity, as supported by our dosimetric findings (**Fig. 5**). It also enables more accurate 3D dose accumulation by capturing anatomical–delivery interplay[43], which is essential for scenarios such as GI re-irradiation with limited cumulative dose tolerance[44].

Moreover, high–temporal resolution 3D imaging enables comprehensive assessment of external–internal motion correlations beyond prior 2D analysis[45]. Our results reveal nonlinear, subject-specific relationships among SI-projected k-space center, internal, and external signals that cannot be captured with 2D dynamic imaging.

To assess real-time MRgRT deployment, B-FIRE inference latency was evaluated on an NVIDIA H200 cluster across varying cropped input patch sizes. Per-diffusion-step latency ranged from 1.6 $ms$ (2700 $ms$ at $T = 1500$) for $256 \times 256$, 1 $ms$ (1500 $ms$ at $T = 1500$) for $128 \times 128$, 0.4 $ms$ ($600\ ms$ at $T = 1500$) for $64 \times 64$, 0.21 $ms$ ($315\ ms$ at $T = 1500$) for $32 \times 32$ to 0.08 $ms$ ($120\ ms$ at $T = 1500$) for $16 \times 16$. These results highlight that latency scales with input tensor size, impacting real-time feasibility. Under fixed latency constraints (e.g., AAPM TG-76[46],; total system latency $< 500\ ms$), inference can be accelerated by reducing diffusion steps with potential fidelity loss, decreasing patch size with increased GPU parallelization to preserve full FOV, or combining both, depending on image quality requirements and available computational resources.

Despite promising results, several limitations remain. First, GPU memory constraints limited training to 2.5D, potentially affecting volumetric coherence; extending to full 3D is needed to improve through-plane consistency under extreme undersampling. Second, evaluation on a single cohort necessitates multi-site validation to establish generalizability across scanners, protocols, and populations. Lastly, the current study lacks MR-Linac data; validation under MR-Linac-specific imaging conditions remains to be established in future work.

## 5. Conclusion

B-FIRE is a binning-free framework for hyper-accelerated non-Cartesian MRI, using a diffusion-optimized CNN–INR backbone to enforce image- and k-space consistency. It enables high-fidelity, real-time reconstruction down to single-spoke sampling and outperforms NuFFT, CS, and several CNN variants across all accelerations on T1-weighted liver data. Critically, it preserves instantaneous physiological motion, overcoming the temporal averaging of motion-binned methods.

## Reference

1. Nyquist, H. Certain Topics in Telegraph Transmission Theory. *Trans. Am. Inst. Electr. Eng.* **47**, 617–644 (1928).
2. Shannon, C. E. Communication in the Presence of Noise. *Proc. IRE* **37**, 10–21 (1949).
3. Ng, J. *et al.* MRI-LINAC: A transformative technology in radiation oncology. *Front. Oncol.* **13**, (2023).
4. Non-cartesian imaging. in *Advances in Magnetic Resonance Technology and Applications* vol. 6 481–498 (Elsevier, 2022).
5. Wright, K. L., Hamilton, J. I., Griswold, M. A., Gulani, V. & Seiberlich, N. Non-Cartesian parallel imaging reconstruction. *J. Magn. Reson. Imaging* **40**, 1022–1040 (2014).
6. Donoho, D. L. Compressed sensing. *IEEE Trans. Inf. Theory* **52**, 1289–1306 (2006).
7. Xu, D. *et al.* Accelerated Patient-specific Non-Cartesian Magnetic Resonance Imaging Reconstruction Using Implicit Neural Representations. *Int. J. Radiat. Oncol.* S036030162506208X (2025) doi:10.1016/j.ijrobp.2025.08.059.
8. Schlemper, J., Caballero, J., Hajnal, J. V., Price, A. N. & Rueckert, D. A Deep Cascade of Convolutional Neural Networks for Dynamic MR Image Reconstruction. *IEEE Trans. Med. Imaging* **37**, 491–503 (2018).
9. Xu, D., Liu, H., Ruan, D. & Sheng, K. Learning Dynamic MRI Reconstruction with Convolutional Network Assisted Reconstruction Swin Transformer. in *Medical Image Computing and Computer Assisted Intervention – MICCAI 2023 Workshops* (eds Woo, J. et al.) vol. 14394 3–13 (Springer Nature Switzerland, Cham, 2023).

10. Xu, D. *et al.* Paired conditional generative adversarial network for highly accelerated liver 4D MRI. *Phys. Med. Biol.* **69**, 125029 (2024).

11. Xu, D. *et al.* Rapid reconstruction of extremely accelerated liver 4D MRI via chained iterative refinement. in *Medical Imaging 2025: Image Processing* (eds Colliot, O. & Mitra, J.) 34 (SPIE, San Diego, United States, 2025). doi:10.1117/12.3034640.

12. Sarma, M. *et al.* Accelerating Dynamic Magnetic Resonance Imaging (MRI) for Lung Tumor Tracking Based on Low-Rank Decomposition in the Spatial–Temporal Domain: A Feasibility Study Based on Simulation and Preliminary Prospective Undersampled MRI. *Int. J. Radiat. Oncol.* **88**, 723–731 (2014).

13. Zhao, N., O'Connor, D., Basarab, A., Ruan, D. & Sheng, K. Motion Compensated Dynamic MRI Reconstruction With Local Affine Optical Flow Estimation. *IEEE Trans. Biomed. Eng.* **66**, 3050–3059 (2019).

14. Otazo, R., Candès, E. & Sodickson, D. K. Low-rank plus sparse matrix decomposition for accelerated dynamic MRI with separation of background and dynamic components: L+S Reconstruction. *Magn. Reson. Med.* **73**, 1125–1136 (2015).

15. Ma, D. *et al.* Magnetic resonance fingerprinting. *Nature* **495**, 187–192 (2013).

16. Lingala, S. G., Hu, Y., DiBella, E. & Jacob, M. Accelerated dynamic MRI exploiting sparsity and low-rank structure: k-t SLR. *IEEE Trans. Med. Imaging* **30**, 1042–1054 (2011).

17. Knoll, F. *et al.* Deep-Learning Methods for Parallel Magnetic Resonance Imaging Reconstruction: A Survey of the Current Approaches, Trends, and Issues. *IEEE Signal Process. Mag.* **37**, 128–140 (2020).

18. Cruz, G., Atkinson, D., Buerger, C., Schaeffter, T. & Prieto, C. Accelerated motion corrected three-dimensional abdominal MRI using total variation regularized SENSE reconstruction. *Magn. Reson. Med.* **75**, 1484–1498 (2016).

19. Holtackers, R. J. & Stuber, M. Free-Running Cardiac and Respiratory Motion-Resolved Imaging: A Paradigm Shift for Managing Motion in Cardiac MRI? *Diagnostics* **14**, 1946 (2024).

20. Uecker, M. *et al.* Real-time MRI at a resolution of 20 ms. *NMR Biomed.* **23**, 986–994 (2010).

21. Goodfellow, I. *et al.* Generative adversarial networks. *Commun. ACM* **63**, 139–144 (2020).

22. Saharia, C. *et al.* Image Super-Resolution Via Iterative Refinement. *IEEE Trans. Pattern Anal. Mach. Intell.* 1–14 (2022) doi:10.1109/TPAMI.2022.3204461.

23. Ronneberger, O., Fischer, P. & Brox, T. U-Net: Convolutional Networks for Biomedical Image Segmentation. in *Medical Image Computing and Computer-Assisted Intervention – MICCAI 2015* (eds Navab, N., Hornegger, J., Wells, W. M. & Frangi, A. F.) vol. 9351 234–241 (Springer International Publishing, Cham, 2015).

24. Larson, A. C. *et al.* Preliminary investigation of respiratory self-gating for free-breathing segmented cine MRI. *Magn. Reson. Med.* **53**, 159–168 (2005).

25. Feng, L. *et al.* Golden-angle radial sparse parallel MRI: Combination of compressed sensing, parallel imaging, and golden-angle radial sampling for fast and flexible dynamic volumetric MRI. *Magn. Reson. Med.* **72**, 707–717 (2014).

26. Segars, W. P., Mahesh, M., Beck, T. J., Frey, E. C. & Tsui, B. M. W. Realistic CT simulation using the 4D XCAT phantom. *Med. Phys.* **35**, 3800–3808 (2008).

27. He, K., Zhang, X., Ren, S. & Sun, J. Deep Residual Learning for Image Recognition. Preprint at https://doi.org/10.48550/arXiv.1512.03385 (2015).

28. Liu, X. *et al.* Implicit Diffusion Models for Continuous Super-Resolution. *Int. J. Comput. Vis.* **133**, 6535–6557 (2025).

29. Lim, B., Son, S., Kim, H., Nah, S. & Lee, K. M. Enhanced Deep Residual Networks for Single Image Super-Resolution. Preprint at https://doi.org/10.48550/arXiv.1707.02921 (2017).

30. Liu, Q. H. & Nguyen, N. An accurate algorithm for nonuniform fast Fourier transforms (NUFFT's). *IEEE Microw. Guid. Wave Lett.* **8**, 18–20 (1998).

31. Feng, L. *et al.* XD-GRASP: Golden-angle radial MRI with reconstruction of extra motion-state dimensions using compressed sensing. *Magn. Reson. Med.* **75**, 775–788 (2016).

32. Xu, D., Descovich, M., Liu, H. & Sheng, K. Robust localization of poorly visible tumor in fiducial free stereotactic body radiation therapy. *Radiother. Oncol.* **200**, 110514 (2024).

33. Xu, D. *et al.* Mask R-CNN assisted 2.5D object detection pipeline of 68Ga-PSMA-11 PET/CT-positive metastatic pelvic lymph node after radical prostatectomy from solely CT imaging. *Sci. Rep.* **13**, 1696 (2023).

34. Oliver, P. A. K. *et al.* Influence of intra- and interfraction motion on planning target volume margin in liver stereotactic body radiation therapy using breath hold. *Adv. Radiat. Oncol.* **6**, 100610 (2021).

35. Peltenburg, J. E. *et al.* Liver Metastases Treated With Magnetic Resonance Imaging Guided Stereotactic Body Radiation Therapy: Outcomes of Tolerability, Acute Toxicity, and Quality of Life From the MOMENTUM Study. *Pract. Radiat. Oncol.* **16**, 48–57 (2026).
36. Moreno-Olmedo, E. *et al.* Daily adaptive MR-guided stereotactic ablative re-irradiation (re-SABR) in oligometastatic liver disease: A single-institution retrospective analysis. *Clin. Transl. Radiat. Oncol.* **56**, 101082 (2026).
37. Stick, L. B., Vogelius, I. R., Risum, S. & Josipovic, M. Intrafractional fiducial marker position variations in stereotactic liver radiotherapy during voluntary deep inspiration breath-hold. *Br. J. Radiol.* **93**, 20200859 (2020).
38. Ehrbar, S. *et al.* Intra- and inter-fraction breath-hold variations and margins for radiotherapy of abdominal targets. *Phys. Imaging Radiat. Oncol.* **28**, 100509 (2023).
39. Sung, J., Choi, Y., Kim, J., Kim, J. W. & Kim, J. Development of in-house software to process real-time cine magnetic resonance images acquired during 1.5 T MR-guided radiation therapy. *Sci. Rep.* **15**, 29515 (2025).
40. Lewis, B. *et al.* Evaluating motion of pancreatic tumors and anatomical surrogates using cine MRI in 0.35T MRgRT under free breathing conditions. *J. Appl. Clin. Med. Phys.* **24**, e13930 (2023).
41. Kurz, C. *et al.* Medical physics challenges in clinical MR-guided radiotherapy. *Radiat. Oncol. Lond. Engl.* **15**, 93 (2020).
42. Holyoake, D. L. P., Aznar, M., Mukherjee, S., Partridge, M. & Hawkins, M. A. Modelling duodenum radiotherapy toxicity using cohort dose-volume-histogram data. *Radiother. Oncol. J. Eur. Soc. Ther. Radiol. Oncol.* **123**, 431–437 (2017).

43. Edvardsson, A., Nordström, F., Ceberg, C. & Ceberg, S. Motion induced interplay effects for VMAT radiotherapy. *Phys. Med. Biol.* **63**, 085012 (2018).

44. Caravatta, L. *et al.* Role of upper abdominal reirradiation for gastrointestinal malignancies: a systematic review of cumulative dose, toxicity, and outcomes on behalf of the Re-Irradiation Working Group of the Italian Association of Radiotherapy and Clinical Oncology (AIRO). *Strahlenther. Onkol. Organ Dtsch. Rontgengesellschaft Al* **196**, 1–14 (2020).

45. Mao, W., Kim, J. & Chetty, I. J. Association Between Internal Organ/Liver Tumor and External Surface Motion From Cine MR Images on an MRI-Linac. *Front. Oncol.* **12**, 868076 (2022).

46. Kissick, M. W. & Mackie, T. R. Task Group 76 Report on 'The management of respiratory motion in radiation oncology' [Med. Phys. 33, 3874-3900 (2006)]. *Med. Phys.* **36**, 5721–5722 (2009).

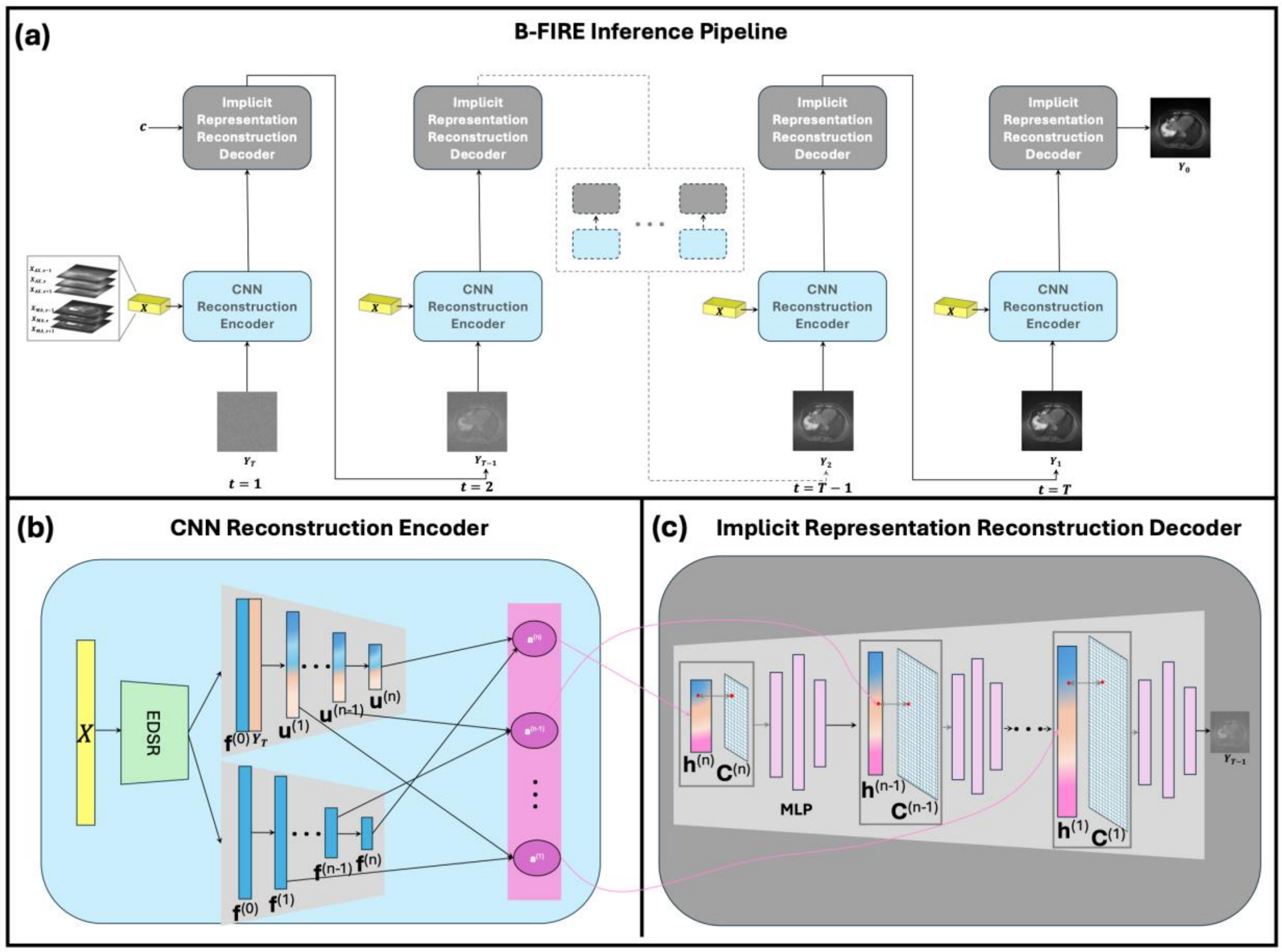

**Fig. 1**: Overview of the B-FIRE framework. (a) The inference process. (b) The CNN reconstruction encoder. (c) The implicit representation reconstruction decoder. $Y_t$, $X$, $f^{(i)}$, $u^{(i)}$, $a^{(i)}$, $h^{(i)}$ and $C^{(i)}$ represent intermediate reconstruction at $t$, model input, hierarchical feature map, intermediate representation, attention-weighted feature, hidden feature, and latent code.

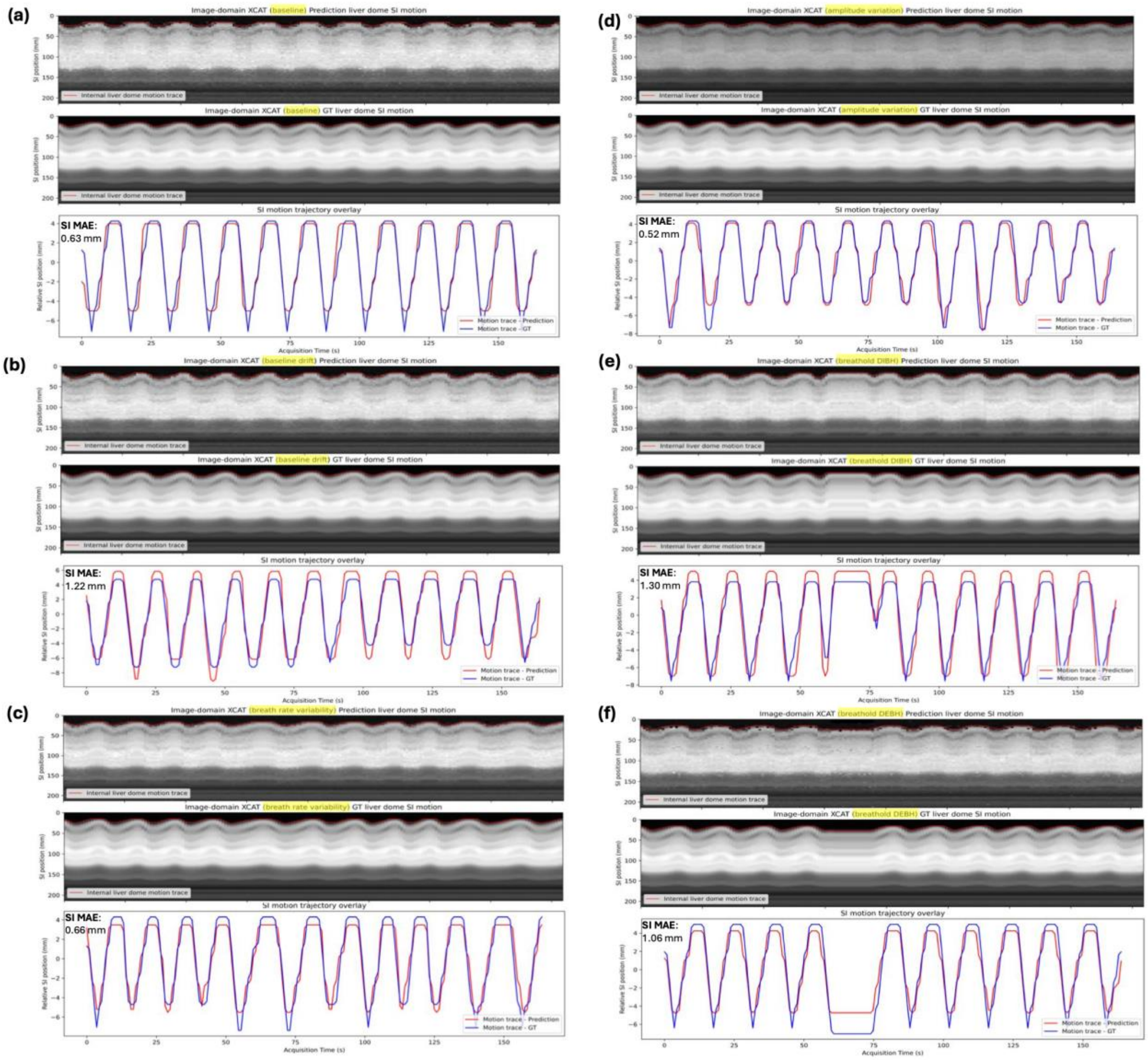

**Fig. 2**: RV8 Binning-free reconstruction using a simulated XCAT phantom under varying breathing conditions. Representative liver dome SI motion traces and GT are shown for baseline, amplitude variation, irregular breathing, and breath-hold (DIBH, DEBH) breathing patterns (a–f). Predicted and GT trajectories are overlaid to assess temporal fidelity, with corresponding MAE reported.

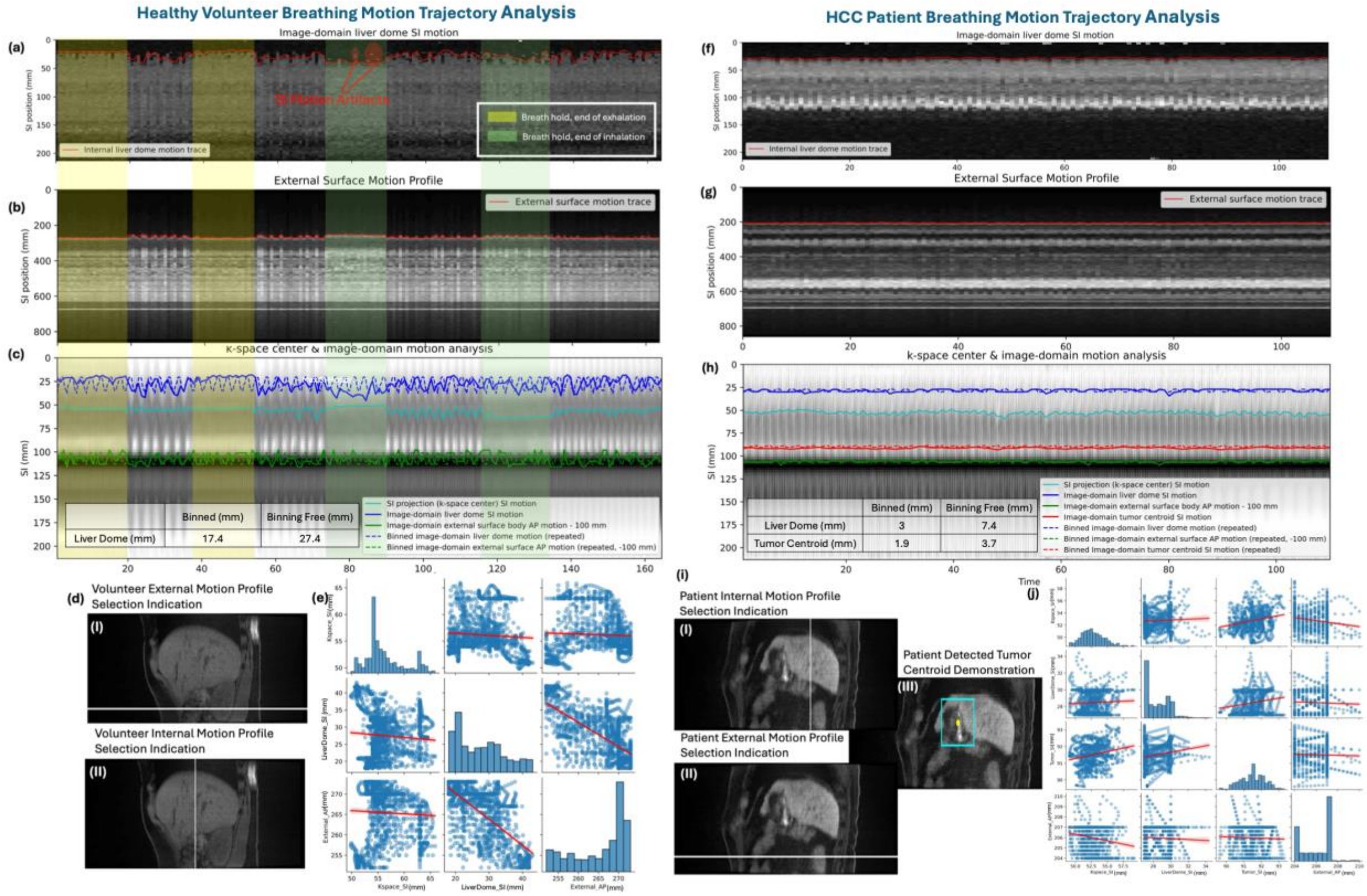

**Fig. 3:** B-FIRE motion trajectory analysis in a healthy volunteer with DIBH, DEBH, and free breathing. Binning-free (RV8) liver dome SI motion (a, f), external surface motion (b, g), and k-space center–projected SI motion with detected trajectories (c, h) are shown, with motion-binned (RV375) references overlaid. Maximum liver dome and tumor centroid motion margins are compared between methods. (d, i) Internal/external motion sampling locations. (e, j) Pair plots show relationships among motion signals, having histograms (diagonal), scatter plots (off diagonal), overlaid with linear regression to indicate correlation strength and direction.

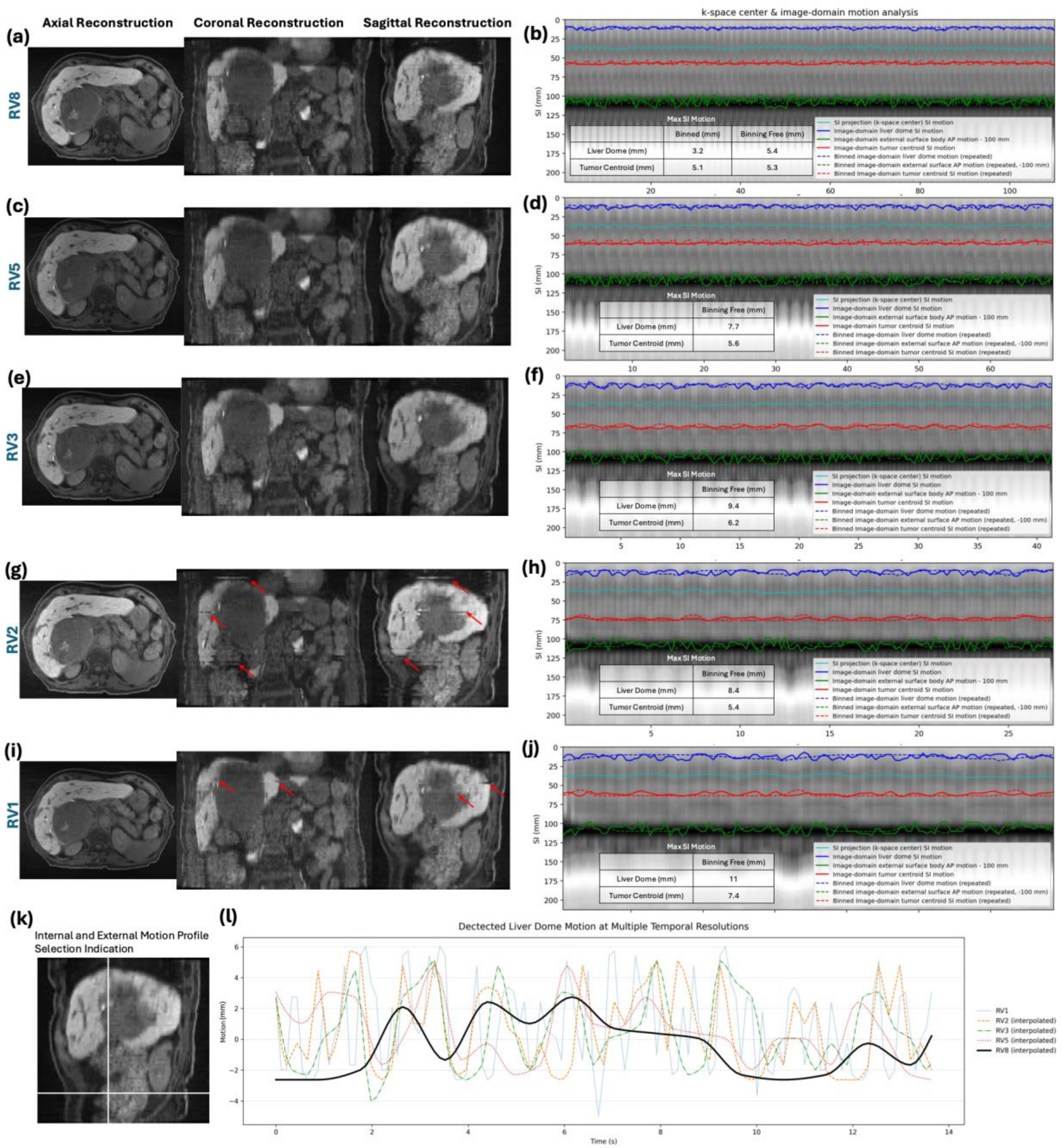

**Fig. 4**: Breathing motion trajectory analysis of binning-free B-FIRE reconstructions across temporal resolutions (RV8–RV1) in a representative free-breathing patient. (a, c, e, g, i): Axial, coronal, and sagittal views with artifacts indicated (red arrows). (b, d, f, h, j): K-space center–projected SI motion with detected trajectories and overlaid motion-binned references. Maximum liver dome and tumor centroid motion margins are compared between methods. (k) Internal and external motion sampling locations. (l) Overlaid liver dome trajectories across resolutions showing consistent motion timing and envelopes.

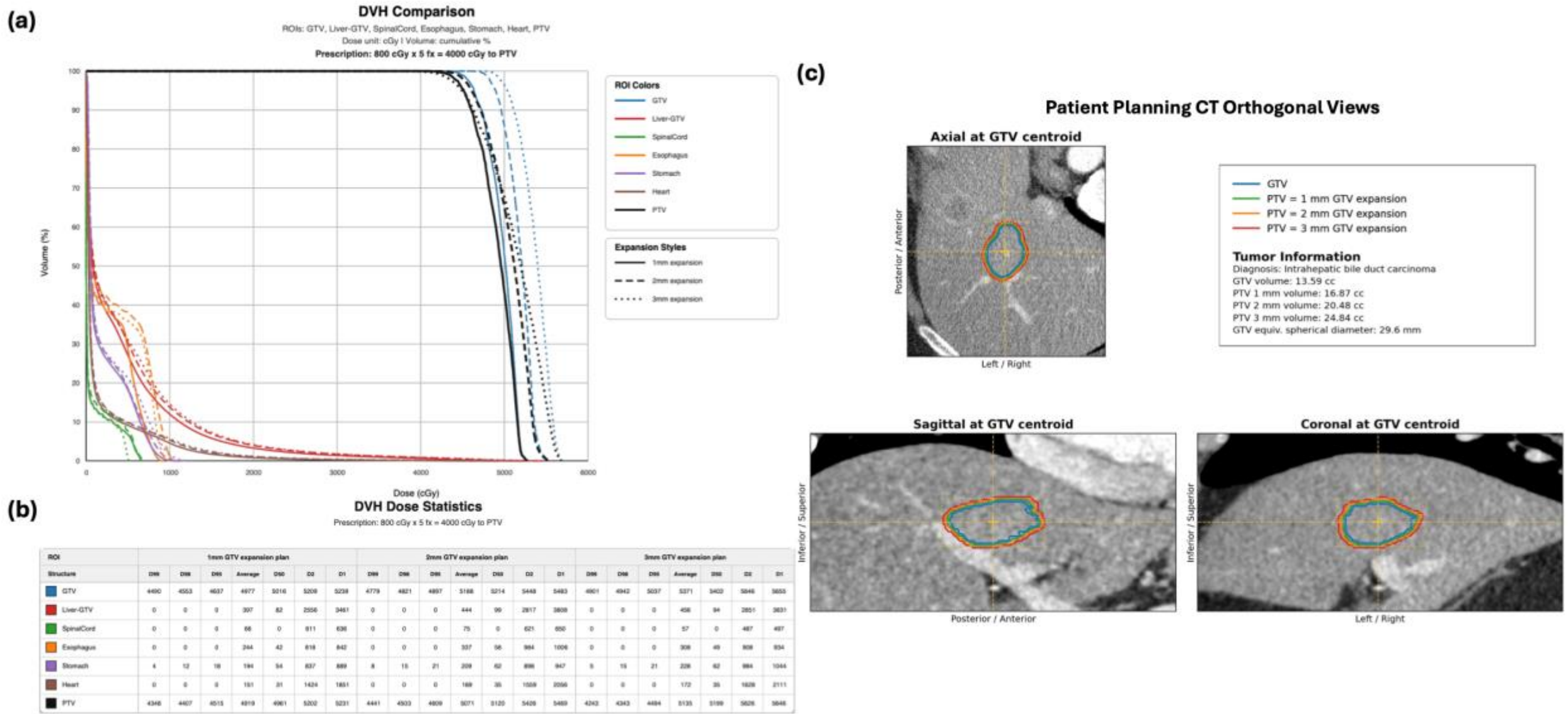

**Fig. 5**: Dosimetric impact of PTV margin selection for a representative liver SBRT patient. (a) DVH comparisons for targets and selected OARs with 1, 2, and 3 mm GTV expansions. (b) Summary of DVH metrics for GTV, PTV, and OARs across margins. (c) Orthogonal planning CT views (axial, sagittal, coronal) centered at the GTV centroid with corresponding GTV/PTV contours.

| Methods | Signal Acquisition | SSIM↑ | RMSE↓ | PSNR (dB) ↑ | DCI↑ |
|---|---|---|---|---|---|
| **B-FIRE†(Proposed)** | RV8 | **0.85 ± 0.01** | **0.025 ± 0.002** | **32.01 ± 0.69** | **0.96 ± 0.01** |
| | RV5 | 0.83 ± 0.01 | 0.028 ± 0.002 | 31.43 ± 0.69 | 0.95 ± 0.01 |
| | RV3 | 0.81 ± 0.02 | 0.032± 0.003 | 30.07 ± 0.78 | 0.95 ± 0.01 |
| | RV2 | 0.79 ± 0.02 | 0.035 ± 0.003 | 29.23 ± 0.78 | 0.93 ± 0.02 |
| | RV1 | 0.78 ± 0.02 | 0.038 ± 0.004 | 28.57 ± 0.87 | 0.91 ± 0.02 |
| **CIRNet** | RV8 | 0.68 ± 0.02 | 0.060 ± 0.006 | 24.52 ± 0.87 | 0.71 ± 0.02 |
| | RV5 | 0.63 ± 0.02 | 0.068 ± 0.007 | 23.30 ± 0.87 | 0.66 ± 0.02 |
| | RV3 | 0.56 ± 0.03 | 0.078 ± 0.009 | 22.27 ± 0.96 | 0.60 ± 0.03 |
| | RV2 | 0.52 ± 0.03 | 0.088 ± 0.010 | 21.13 ± 0.96 | 0.56 ± 0.03 |
| | RV1 | 0.50 ± 0.03 | 0.098 ± 0.012 | 20.20 ± 1.04 | 0.54 ± 0.03 |
| **Re-Con-GAN** | RV8 | 0.64 ± 0.02 | 0.072 ± 0.007 | 22.87 ± 0.87 | 0.51± 0.02 |
| | RV5 | 0.59 ± 0.03 | 0.081 ± 0.009 | 21.84 ± 0.96 | 0.56 ± 0.03 |
| | RV3 | 0.53 ± 0.03 | 0.091 ± 0.010 | 20.82 ± 0.96 | 0.52 ± 0.03 |
| | RV2 | 0.49 ± 0.03 | 0.10 ± 0.012 | 19.80 ± 1.04 | 0.48 ± 0.03 |
| | RV1 | 0.47 ± 0.03 | 0.12 ± 0.014 | 18.86 ± 1.04 | 0.47 ± 0.04 |
| **Compressed Sensing** | RV8 | 0.51 ± 0.03 | 0.12 ± 0.014 | 18.54 ± 1.04 | 0.45 ± 0.03 |
| | RV5 | 0.46 ± 0.03 | 0.14 ± 0.018 | 17.11 ± 1.13 | 0.41 ± 0.03 |
| | RV3 | 0.40 ± 0.04 | 0.16 ± 0.023 | 15.72 ± 1.22 | 0.37 ± 0.04 |
| | RV2 | 0.39 ± 0.05 | 0.19 ± 0.029 | 14.31 ± 1.30 | 0.33 ± 0.04 |
| | RV1 | 0.38 ± 0.05 | 0.23 ± 0.036 | 12.96 ± 1.39 | 0.33± 0.05 |
| **Cascade CNN** | RV8 | 0.60 ± 0.03 | 0.10 ± 0.012 | 20.08 ± 1.04 | 0.42 ± 0.03 |
| | RV5 | 0.54 ± 0.03 | 0.11 ± 0.013 | 19.22 ± 1.04 | 0.39 ± 0.03 |
| | RV3 | 0.47 ± 0.04 | 0.12 ± 0.016 | 18.41 ± 1.13 | 0.36 ± 0.04 |
| | RV2 | 0.46 ± 0.04 | 0.13 ± 0.017 | 17.68 ± 1.13 | 0.32 ± 0.04 |
| | RV1 | 0.45 ± 0.04 | 0.14 ± 0.020 | 17.09 ± 1.22 | 0.31 ± 0.04 |
| **NuFFT** | RV8 | 0.48 ± 0.04 | 0.15 ± 0.020 | 16.31 ± 1.13 | 0.37 ± 0.04 |
| | RV5 | 0.41 ± 0.04 | 0.18 ± 0.025 | 15.04 ± 1.22 | 0.34 ± 0.04 |
| | RV3 | 0.37 ± 0.05 | 0.21 ± 0.031 | 13.72 ± 1.30 | 0.31 ± 0.04 |
| | RV2 | 0.36 ± 0.05 | 0.24 ± 0.039 | 12.34 ± 1.39 | 0.26 ± 0.05 |
| | RV1 | 0.35 ± 0.05 | 0.28 ± 0.051 | 11.01 ± 1.56 | 0.26 ± 0.05 |

† *Statistically significant compared to all baseline methods with* $p < 0.001$ *(Wilcoxon signed-rank test).*

**Table 1**: Quantitative evaluation of B-FIRE and benchmarks on the XCAT phantom test cohort (binning-free simulation with known GT) across acquisition settings. SSIM, PSNR (dB), RMSE, and Detectability consistency index (DCI) are reported; reconstructions are normalized to [0,1], SSIM/PSNR computed in 3D, and best results bolded.